\pdfoutput=1
\documentclass[11pt]{article}

\usepackage{acl}
\usepackage[utf8]{inputenc} 
\usepackage[T1]{fontenc}    
\usepackage{hyperref}       
\usepackage{url}            
\usepackage{booktabs}       
\usepackage{amsfonts}       
\usepackage{nicefrac}       
\usepackage{microtype}      
\usepackage{xcolor}         
\usepackage{natbib}
\usepackage{float}
\usepackage{graphicx} 
\usepackage{epstopdf}
\usepackage{amsmath}
\usepackage{times}
\usepackage{latexsym}
\usepackage{multirow}
\usepackage{multicol}
\usepackage{subfigure}
\usepackage{diagbox}
\usepackage{makecell}
\usepackage{graphics}
\usepackage{amsfonts,amssymb} 
\usepackage{booktabs}
\usepackage{tabularx}
\usepackage{enumitem}
\usepackage[T1]{fontenc}
\usepackage[utf8]{inputenc}
\usepackage{color}
\usepackage[noend]{algpseudocode}
\usepackage{algorithmicx,algorithm}
\usepackage{wrapfig}
\usepackage{pifont}

\definecolor{newgreen}{RGB}{0,102,51}
\title{Improving Temporal Generalization of Pre-trained Language Models with Lexical Semantic Change}

\author{Zhaochen Su$^{1}$, Zecheng Tang$^1$\thanks{\; Zhaochen Su and Zecheng Tang contribute equally.}, Xinyan Guan$^1$, Juntao Li$^{1}$\thanks{\; Juntao Li is the Corresponding Author.}, Lijun Wu$^2$, Min Zhang$^{1}$ \\  
 $^{1}$Institute of Computer Science and Technology, Soochow University, China \\
 $^{2}$Microsoft Research Asia\\
 \texttt{\{suzhaochen0110,zctang2000,guanxy0406\}@gmail.com}; \\
  \texttt{\{ljt,minzhang\}@suda.edu.cn}; 
 \texttt{lijuwu@microsoft.com} \\
 }
 
\begin{document}
\maketitle
\begin{abstract}
Recent research has revealed that neural language models at scale suffer from poor temporal generalization capability, i.e., language model pre-trained on static data from past years performs worse over time on emerging data.  
Existing methods mainly perform continual training to mitigate such a misalignment. While effective to some extent but is far from being addressed on both the language modeling and downstream tasks.
In this paper, we empirically observe that temporal generalization is closely affiliated with lexical semantic change, which is one of the essential phenomena of natural languages.
Based on this observation, we propose a simple yet effective lexical-level masking strategy to post-train a converged language model.
Experiments on two pre-trained language models, two different classification tasks, and four benchmark datasets demonstrate the effectiveness of our proposed method over existing temporal adaptation methods, i.e., continual training with new data.
Our code is available at \url{https://github.com/zhaochen0110/LMLM}.

\end{abstract}

\section{Introduction}

Neural language models (LMs) are one of the frontier research fields of deep learning.
With the explosion of model parameters and data scale, these language models demonstrate superior generalization capability, which can enhance many downstream tasks even under the few-shot and zero-shot settings~\cite{radford2018improving,radford2019language,brown2020language,zhang2021commentary}.
Although these models have achieved remarkable success, they are trapped by the time-agnostic setting in which the model is trained and tested on data with significant time overlap.
However, real-world applications usually adopt language models pre-trained on past data (e.g., BERT \cite{devlin2019bert} and RoBERTa~\cite{liu2019roberta}) to enhance the downstream task-specific models for future data, resulting in a temporal misalignment~\cite{luu2021time}. 
Recent works have empirically demonstrated that such a misalignment hurts the performance of both the upstream language models and downstream task-specific methods~\cite{lazaridou2021mind,rottger2021temporal}.

\begin{figure}[t]
  \centering
  \includegraphics[width=1\linewidth]{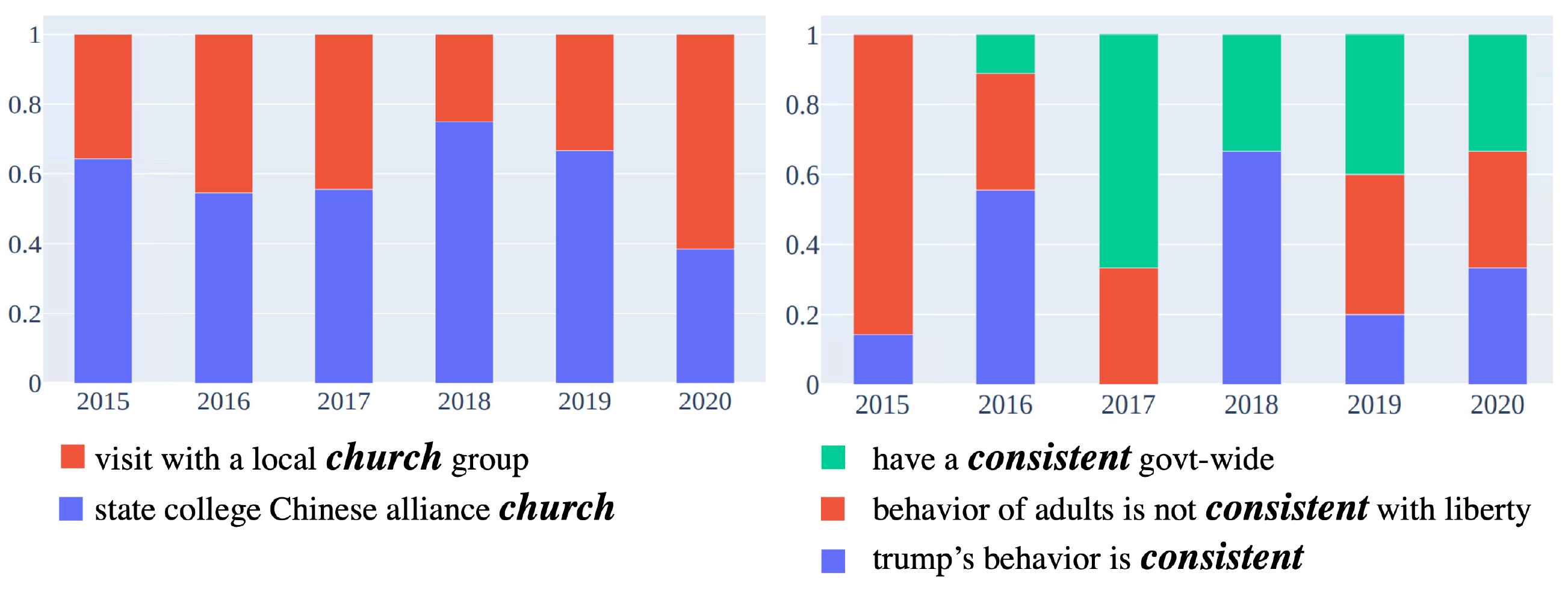}
  \caption{Examples of lexical semantic change across time, where the semantic of word \textit{church} is stable as it often refers to a building or a local congregation for Christian religious activities. However, the semantic of word \textit{consistent} varies dramatically at different times.}
  \label{fig:type distribution}
\end{figure}

To better understand and solve the temporal misalignment problem, a series of studies have been launched on pre-trained language models (PLMs) and downstream tasks.
The analysis on PLMs~\cite{lazaridou2021mind} revealed that PLMs (even with larger model sizes) encounter a serious temporal generalization problem, and the misalignment degree increases with time.
They also found that continually pre-training PLMs with up-to-the-minute data does mitigate the temporal misalignment problem but suffers from catastrophic forgetting and massive computational cost since further pre-training the converged PLMs is as difficult as pre-training from scratch.
The study on downstream tasks further indicates that temporal adaptation (i.e., continually pre-training with unlabelled data that is mostly overlapped in time), while effective, has no apparent advantages over domain adaptation~\cite{rottger2021temporal} (i.e., continually pre-training with domain-specific unlabelled data) and fine-tuning on task-specific data from the target time~\cite{luu2021time}.

To analyze the reason behind the limited performance of temporal adaptation, we launch a study from the lexical level, which also matches the token-level masking operation in advanced PLMs.
Unlike existing research that launches analysis on part-of-speech (POS), topic words, and newly emerging words,
we mainly explore the correlations between language model performance and tokens/words with salient lexical semantic change, which is also an extensively studied concept in computational linguistics~\cite{dubossarsky2015bottom,hamilton2016diachronic,giulianelli2020analysing} to investigate how the semantics of words change over time.
Experimental results demonstrate that tokens/words with salient lexical semantic change do contribute much more than the rest of tokens/words to the temporal misalignment problem, manifested as their significantly higher perplexity~(\textit{ppl.}) over randomly sampled tokens from the target time.
However, the widely-adopted masked language model (MLM) objective in state-of-the-art PLMs uniformly deals with each token/word, letting the salient lexical-level semantic change information over time being overwhelmed by other tokens/words, which can also explain why temporal adaptation has no obvious advantage compared with domain adaptation.

Based on the above findings, we propose a lexical-based masked Language Model (LMLM) objective to capture the lexical semantic change between different temporal splits.
Experimental results demonstrate that our proposed method yields significant performance improvement over domain adaptation methods on two different PLMs and four benchmark datasets. 
Extensive studies also show that LMLM is effective when utilizing different lexical semantic change metrics.

In a nutshell, our contributions are shown below:
\begin{itemize}[leftmargin=*]
\setlength{\itemsep}{0pt}
\item We empirically study the temporal misalignment of PLMs at the lexical level and reveal that the tokens/words with salient lexical semantic change contribute much more to the misalignment problem than other tokens/words. We also disclose that such lexical temporal misalignment information can be overwhelmed by the masked language model training objective of PLMs, resulting in limited performance improvement over temporal and domain adaptation methods.
\item We propose a simple yet effective Lexical-based Masked Language Model (LMLM) objective to improve the temporal generalization of PLMs.
\item Experiments on two PLMs and four different benchmark datasets confirm that our proposed method is extensively effective in addressing the temporal misalignment problem for downstream tasks, which can significantly outperform existing temporal and domain adaptation methods.
\end{itemize}

\begin{figure*}[t]
    \centering
    \includegraphics[width=1\textwidth]{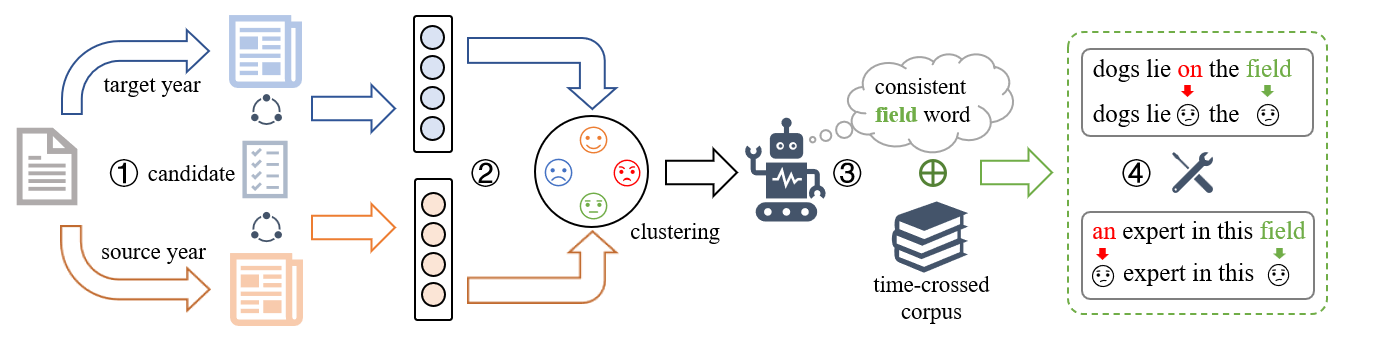}
    \caption{The pipeline of detecting semantic change words (\ding{172}$\sim$~\ding{174}) and the Lexical-based Masked Language Model (LMLM) objective (\ding{175}). In step \ding{175}, the words colored with red are randomly sampled and the words with salient semantic change are colored with green. ALL colored words/tokens are masked during the pre-training stage.}
    \label{fig:pipeline}
\end{figure*}

\section{Linking Temporal Misalignment with Lexical Semantic Change}
Recent work on temporal adaptation~\cite{rottger2021temporal} has found that post-tuning the converged PLMs with unlabeled time-specific data by reusing the MLM objective can make the PLMs perceive related event-driven changes in language usage.
Such adaptation can achieve decent performance because the widely-adopted MLM objective can capture the overall changes in the data distribution by randomly masking a specific ratio of the whole sequence. 
However, such a training objective makes the lexical-level temporal information ignored or overwhelmed by the time-agnostic tokens/words, resulting in little to no performance superiority over domain adaptation methods.
Based on the above background, it is natural to explore the role of lexical-level temporal information in temporal adaptation, i.e., whether these tokens/words with salient lexical-semantic changes\footnote{The concept of semantic change is also essential in computational linguistics~\cite{gulordava2011distributional,bamler2017dynamic,rosenfeld2018deep,del2018short,giulianelli2020analysing}.} over time impair the temporal adaptation performance. 
As a result, we launch a thorough study from the perspective of lexical semantic change to figure out the reason behind the limited performance of temporal adaptation in the specific domain.
To the best of our knowledge, this is the first study that explores the correlation between the lexical-semantic change and the temporal adaptation of PLMs.
We will firstly illustrate our methods to find those semantic changed words in Section~\ref{sec:dectection} and introduce the discovery experiment as well as analyze the results in Section~\ref{sec:lsc_ana}.


\subsection{Lexical Semantic Change Detection}
\label{sec:dectection}
To obtain the semantic changed words, we design a lexical semantic change detection process.
For better illustration, we decompose the process into three steps: candidate words selection, feature extraction \& Clustering, and semantic change quantification, which are correspond with the step~\ding{172}$\sim$~\ding{174} in Figure~\ref{fig:pipeline}.

\paragraph{Candidate Words Selection}
Before obtaining the representation of each word, we sample a certain number of candidate words $\mathcal{W}^{t} = \{w_{1}^{t}, w_{2}^{t}, \cdots, w_{k}^{t}\}$ from the texts $\mathcal{D}^{t}$ of time $t$.
Considering that different texts have different domains (politics, culture, history), most keyword extraction methods either heavily rely on dictionaries and a fussy training  process~\cite{witten2005kea,rose2010automatic} or are too simple to handle such intricate domain changes, i.e., TF-IDF~\cite{ramos2003using}.
Instead, we turn to YAKE!~\cite{campos2018yake}, a feature-based and unsupervised system to extract keywords in one document.
Since the goal is to measure the lexical semantic change among different time splits, we further filter the $\mathcal{W}^{t}$ by calculating the number of the candidate words in different periods and removing the words that are repetitive, too few, or have no real meanings, e.g., pronouns, particles, mood words, etc.

\paragraph{Feature Extraction and Clustering}
Given a word $w_{i}$ and one text $d_{i}^{t}=(t_{1},\cdots,t_{i},\cdots,t_{n})$, where $d_{i}^{t}\in \mathcal{D}^{t}$ and $t_{i}=w_{i}$, we utilize a pre-trained language model BERT~\cite{devlin2019bert} to contextualise each text as the representation $r_{i}^{t}$.
Specifically, we look up the sentences in $\mathcal{D}^{t}$ which contain the same candidate words in $\mathcal{W}^{t}$ and feed them into BERT to extract the corresponding word representations followed by aggregating them together~\cite{giulianelli2020analysing}.
It is worth noting that we extract the representations from the last layer of the BERT model in all experiments, but we also consider extracting the features from the shallow layers of the BERT model. More details can be referred to in Appendix~\ref{appdix:upper_layer}.  

To prevent too much information brought by the long sequences overwhelming the meaning of the candidate words, we specify 128 as the size of occurrence window around the word $w_{i}$, i.e., truncating the redundant part of each sentence.
After obtaining $N$ usage representations for each word, we combine them together as representation matrix $\mathcal{R}_{i}^{t}=(r_{1}^{t},r_{2}^{t},\cdots ,r_{N}^{t})$ and normalise it.

To distinguish the different semantic representations of each word, we utilize the $K$-Means algorithm, which can automatically cluster the similar word usage type into $K$ groups after $p$ turns according to the representation matrix of each word.
Details about the $K$-Means algorithm is elaborated in Appendix~\ref{appdix:k_means}.
After clustering, we count the number of sentences in each cluster and calculate the frequency distribution for the candidate word $w_{i}$.
When normalized, the frequency distribution can be viewed as the probability distribution $p^{t}_{i}$ over usage types for the candidate word $w_{i}$ at the time $t$.
To meet our temporal settings, we should get the probability distributions for the same candidate word in different periods for comparison.

\paragraph{Semantic Change Quantification}
To measure the difference between the probability distributions $p^{t}_{i}$ and $p^{t^{\prime}}_{i}$ of the same candidate words in different periods over word usages, we utilize the Jensen-Shannon divergence~\cite{lin1991divergence} metric:
\begin{equation}
\begin{aligned}
    {\rm JSD}(p^{t}_{i}, p^{t^{\prime}}_{i})& =  \mathbb{H}\left[\frac{1}{2}(p^{t}_{i}+p^{t^{\prime}}_{i})\right] \\
    &-\frac{1}{2}\left[\mathbb{H}(p^{t}_{i})-\mathbb{H}(p^{t^{\prime}}_{i})\right],
\end{aligned}
\end{equation}
where $\mathbb{H}$ is the Boltzmann-Gibbs-Shannon entropy~\cite{ochs1976basic}.
High JSD represents the different frequency distributions, i.e., significant lexical semantic change of the word $t_{i}$, and visa versa.
We utilize a hyper-parameter $k$ to control the degree of the lexical semantic change. 
Specifically, we rank the candidate words according to their JSD values and sample the top-$k$ words as the salient semantic changed words.
Several other metrics can also quantify the lexical semantic change, e.g., Entropy Difference (ED)~\cite{nardone2014entropy} and Average pairwise distance (APD)~\cite{bohonak2002ibd}, and we will compare the performance among them below.

\subsection{Discovery Experiment \& Analysis}
\label{sec:lsc_ana}
To highlight the influence of the salient semantic changed words, we design a special masked language modeling objective LMLM, which first masks the candidate words $\mathcal{W}^{t}$ in the texts.
Details of the LMLM objective are elaborated in section~\ref{sec:LMLM}.
All the experiments in this section are conducted with the ARXIV dataset\footnote{\url{https://arxiv.org/help/oa/index}}, which contains the abstracts of five subjects in different periods, e.g., CS, Math, etc.
We apply the pre-trained BERT-base model\footnote{\url{https://github.com/google-research/bert}} which has been pre-trained on a large corpus and evaluate it with the latter-released testing sets\footnote{The BERT model is pre-trained with the data in 2015, while the three testing sets are after 2017.} by reporting the Perplexity (\textit{ppl.}) value.
All the above data are tokenized with Moses\footnote{\url{ https://github.com/alvations/sacremoses}}, and non-English documents are removed.

\begin{figure}[t]
\centering
\label{fig:prob}
\subfigure[]{
\begin{minipage}[t]{0.225\textwidth}
  \includegraphics[width=1\linewidth]{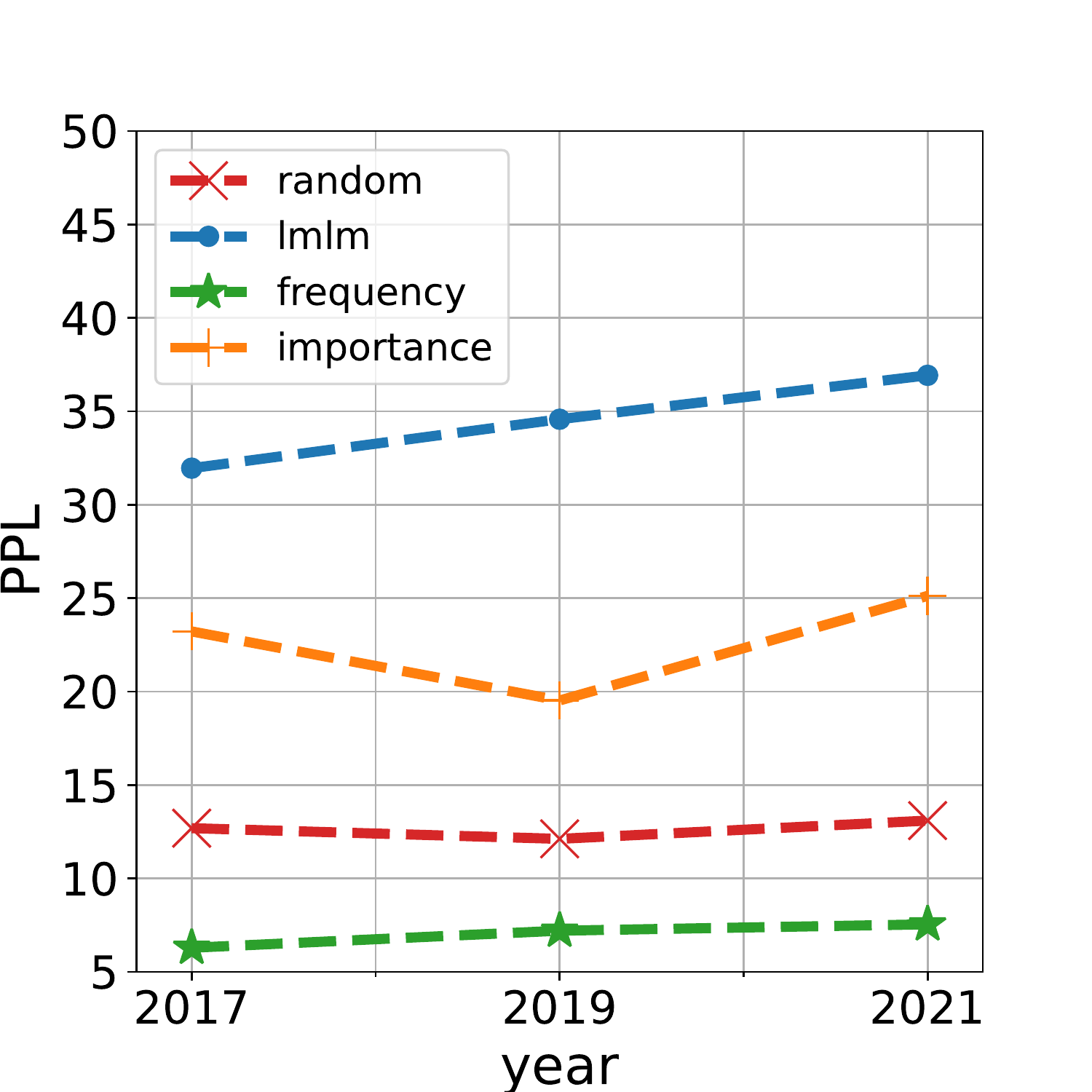}
  \label{fig:sub1}
\end{minipage}
}
\subfigure[]{
\begin{minipage}[t]{0.225\textwidth}
  \includegraphics[width=1\linewidth]{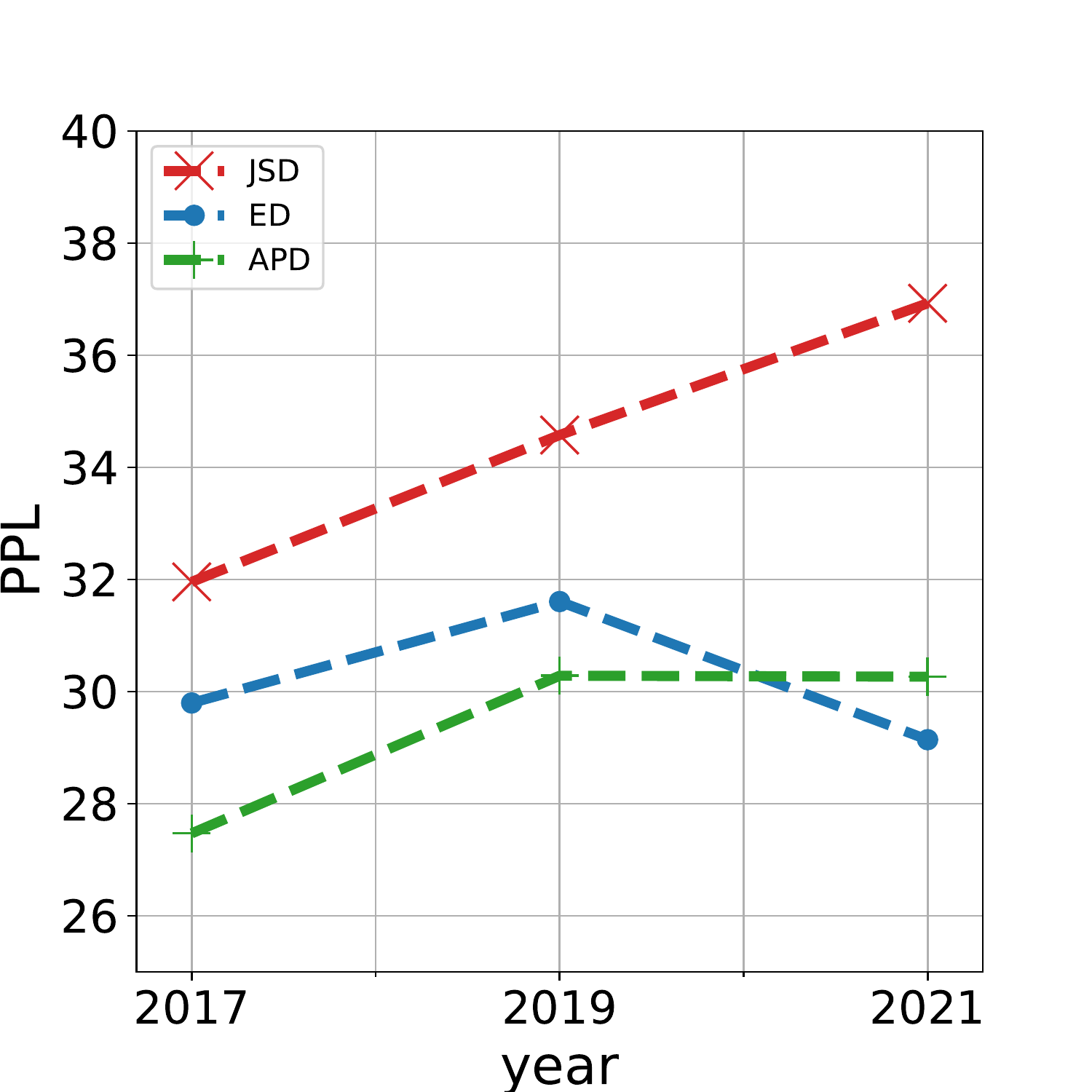}
  \label{fig:sub2}
\end{minipage}
}
\caption{Results of the \textit{ppl.}~value. Figure~(a) and (b) shows the effect of the semantic changed words and the results of different quantification metrics respectively.}
\end{figure}

\paragraph{Influence of the Semantic Changed Tokens}
For comparison, we introduce four masking strategies: random masking, frequency masking, importance masking, and LMLM.
The masking ratio of the strategies above is 15\%.
The random masking strategy, as mentioned above, masks the tokens in the texts randomly, while the frequency masking strategy masks the tokens according to the lexical occurrence frequency, and the importance masking strategy masks the tokes according to the YAKE! scores.
Details of the masking strategies are illustrated in Appendix~\ref{appdix:semantic}.
The results are shown in the figure~\ref{fig:sub1}.
We can observe that the \textit{ppl.} of the LMLM (blue dotted curve) is much higher than the others, which indicates that it is hard for the PLM to predict the lexical semantic changed tokens. 
The rising trend of four curves shows that the PLM performs increasingly worse when predicting future utterances further away from their training period.

\paragraph{Influence of the Quantification Metric}
Furthermore, We utilize three current popular metrics: Jensen-Shannon divergence (JSD),  Entropy Difference (ED), and Average Pairwise Distance (APD) to measure the semantic change.
The results are shown in the figure~\ref{fig:sub2}.
Since the slope of the red curve (JSD) is much higher than the others, which means the candidate words selected with the JSD metric are hard to predict, i.e., the semantic change phenomenon of those words is more significant, we apply the JSD metric in the later experiments to find the candidate words $\mathcal{W}^{t}$.

\section{LMLM Objective}
\label{sec:LMLM}
Masked Language Model (MLM) objective is a widely-adopted unsupervised task proposed by BERT~\cite{devlin2019bert}, which randomly masks proportional tokens and predicts them. 
Since the degradation of the PLM in a specific domain over time is mainly attributed to the words with salient semantic change, we should make the PLM more aware of them.
Thus, we propose our Lexical-based Masked Language Model (LMLM) objective.
Contrary to the traditional random masking strategy (MLM), LMLM preferentially masks the words with salient semantic change over time.
Formally, given the text set $\mathcal{D}^{t}=\{d_{1}^{t}, d_{2}^{t}, \cdots, d_{n}^{t}\}$ at time $t$, we select the candidate words $\mathcal{W}^{t}$ with the aforementioned detection method and rank them according to their JSD value.
Then, we select $k~(k \in \{100, 200, \cdots, 1000\})$ words which have relative high scores as the masking candidates $\mathcal{W}_{mask}^{t}$.
Given masking ratio $\alpha$, LMLM firstly selects the words in the $\mathcal{W}_{mask}^{t}$ to mask.
If there are not enough candidates to meet the total number of masking tokens, LMLM masks the other words in the text randomly.
The whole process is corresponding to the step \ding{175} in Figure~\ref{fig:pipeline}.
Assuming it masks $m$ tokens in total and the sentence after masking is $d_{i}^{t^{\prime}}$.
The optimization objective of LMLM can be formulated as:
\begin{equation}
\mathcal{L}_{\mathrm{LMLM}} = - \sum\limits_{j=1}^{m} \log P(x=w_{j}|d_{i}^{t^{\prime}};\theta).
\label{bert_mlm}
\end{equation}
\section{Experiments}
We conduct experiments on the classification task by employing the pre-trained BERT model implemented with the Hugging-Face transformers package\footnote{\url{https://huggingface.co}} in all experiments.
Further details about model training and parameters can be found in Appendix~\ref{appdix:para}.
We will introduce the datasets and the time-stratified settings in Section~\ref{sec:settings}, the baselines in Section~\ref{sec:baseline}, and show the results in Section~\ref{sec:res}.

\begin{table}[t]
\centering
\small
\begin{tabular}{c|c|c|l}
\toprule
Dataset \quad & \quad Usage \quad & Time \quad & \#Sentences \quad  \\ 
\midrule
\multirow{2}{*}{ARXIV} & Fine. & 2007$\sim$2019~$\dagger$ & 160,000 \\
                       & Pre.& 2021 & 3,800,000 \\
\midrule
\multirow{2}{*}{PoliAff} & Fine. & 2015, 2016 & 10,000 \\
                         & Pre.~$\ddagger$  & 2017 & 1,000,000 \\
\midrule
\multirow{3}{*}{RTC} & Fine. & Apr. 2017 & 20,000 \\
                     & Pre. & Apr. 2018 & 2,000,000 \\
                     & Pre. & Aug. 2019 & 2,000,000 \\
\bottomrule
\end{tabular}
\caption{Statistics of the datasets, where the time splits of the ARIXV fine-tuning data (marked with $\dagger$) is on a four-year cycle, and the pre-training data of PoliAff dataset (marked with $\ddagger$) is WMT17.}
\label{tab:data}
\end{table}

\subsection{Basic Settings}
\label{sec:settings}
\paragraph{Datasets}
To ensure the PLM is trained with the data in the specific domain, we select the data with the same or similar distributions between the upstream and downstream stages.
We choose the ARXIV dataset for the scientific domain and Reddit Time Corpus (RTC) dataset\footnote{\url{https://github.com/paul-rottger/temporal-adaptation}} for the political domain.
We also turn to two different datasets with a similar distribution for pre-training and fine-tuning, respectively.
Specifically, we select WMT News Crawl (WMT)\footnote{\url{https://data.statmt.org/news-crawl/}} dataset, which contains news covering various topics, e.g., finance, politics, etc, as unlabeled data and PoliAff\footnote{\url{https://github.com/Kel-Lu/time-waits-for-no-one}} dataset  in politic domain as labeled data.

\paragraph{Time-Stratified Settings}
Generally, the PLM is adapted to temporality using unlabelled data, fine-tuned with the downstream labeled data, and then evaluated with the testing data which has the same time as the pre-training data.
We set the $k$ as 500 in all experiments.
As for the ARXIV dataset, we utilize the unlabeled data in 2021 for pre-training and extract five years of data from 2011 to 2019 on a four-year cycle for fine-tuning as well as the data in 2021 for testing.
Similarly, we collect the data in 2015 and 2016 from the PoliAff dataset as the fine-tuning data and test the model with the data in 2017.
For the RTC dataset, we follow the previous work~\cite{rottger2021temporal} to select the unlabeled News Comments dataset for post-training and the political subreddit subset for fine-tuning.
However, the number of masking candidates $k$ is less than 500 in most RTC fine-tuning sets of different time splits, which could make the LMLM strategy be regarded as the random masking strategy. 
Thus, we select the data in April 2017 for fine-tuning (where $k \geq 500$ in this subset) and the data in April 2018 and August 2019 for testing.
Detailed data statistics are shown in Table~\ref{tab:data}.

\subsection{Baselines}
\label{sec:baseline}
To meet the time-stratified settings, we select the temporal adaptation \textbf{TAda} method~\cite{rottger2021temporal} as baseline, which first incorporates the temporal information into the PLM by utilizing the time-specific unlabeled data for pre-training and then adapt the PLM to the downstream task with the supervised data.
Besides the temporal adaptation method, we also turn to some up-to-date domain adaptation methods since previous work~\cite{rottger2021temporal} points out that such method can mitigate the temporal misalignment problems to some extent.
Specifically, we select PERL~\cite{ben2020perl} and DILBERT~\cite{lekhtman2021dilbert} methods, and implement them under the time-stratified settings.
Details of the domain adaptation methods are shown in the Appendix~\ref{appdix:methods}.
We calculate the F1 score as the testing results for all the experiments.

\begin{table}[t]   
\centering
\resizebox{\columnwidth}{!}{
\begin{tabular}{l|cccc|c}
\toprule
\multirow{2}{*}{Method} & \multicolumn{5}{c}{Fine-tuning Data} \\
 \cmidrule{2-6}
 & 2007 & 2011 & 2015 & 2019 & Avg. \\ 
\midrule
 TAda  & 82.97 & 84.72 & 84.82 & 84.99 & 84.38 \\
 + PERL & 75.67 & 79.20 & 78.89 & 78.77 & 78.13 \\ 
 + DILBERT & 82.62 & 83.89 & 84.04 & 84.22 & 83.69 \\
 + LMLM & \textbf{84.93} & \textbf{86.52} & \textbf{86.49} & \textbf{87.22} & \textbf{86.29} \\
\bottomrule
\end{tabular}}
\caption{Results of the ARXIV dataset. }
\label{tabel:arxiv-main-result}
\end{table}
\begin{table}[t]
\centering
\small
\begin{tabular}{l|cc|c}
\toprule
\multirow{2}{*}{Method} & \multicolumn{3}{c}{Fine-tuning Data}  \\
\cmidrule{2-4}
& \quad 2015 \quad & 2016 \quad\quad &\quad Avg. \quad\quad  \\
\midrule
TAda      & \quad 66.05 \quad    & 72.94\quad\quad &\quad 69.50 \quad\quad \\
+ PERL    & \quad 61.79 \quad    & 68.21\quad\quad &\quad 65.00 \quad\quad \\
+ DILBERT & \quad 63.89 \quad    & 69.86\quad\quad &\quad 66.88 \quad\quad \\
+ LMLM    & \quad\bf 67.00 \quad & \bf 74.10\quad\quad &\quad \bf 70.55 \quad\quad \\
                            
\bottomrule
\end{tabular}
\caption{Results of the PoliAff dataset.}
\label{poli}
\end{table}


\subsection{Main Results}
\label{sec:res}
\paragraph{ARXIV Dataset}
The results of the ARXIV dataset are shown in Table~\ref{tabel:arxiv-main-result}, and we can observe that applying domain adaptation methods under the time-stratified settings aggravate the temporal misalignment problem as the scores of the PERL and DILBERT methods are not as high as those for utilizing the TAda directly.
However, the performance of LMLM is much better than the other three methods.

\paragraph{PoliAff Dataset}
We report the results of the PoliAff dataset in Table~\ref{poli}. 
Although there is a slight domain difference between the pre-training and fine-tuning data, i.e., news and politic, the LMLM can still achieve the best results, and the domain adaptation methods still perform worse than the temporal adaptation methods.

\paragraph{RTC Dataset}
The results of the RTC dataset is shown in Table~\ref{tabel:rtc-main-result}, and we can find the similar tendency as the previous results, i.e., LMLM still achieve the best performance.
However, the differences among the four methods in RTC dataset are much smaller compared with the previous results, which is largely due to the slight dynamic temporality of the RTC dataset.

\begin{table}[t]
\centering
\small
\begin{tabular}{l|cc|c}
\toprule
\small
\multirow{2}{*}{Method} & \multicolumn{3}{c}{Testing Data} \\
\cmidrule{2-4} 
    & Apr.~2018 & Aug.~2019 & \quad Avg. \quad\quad  \\
\midrule
TAda   & 41.78 & 38.14 & \quad 39.96 \quad\quad  \\
+ PERL & 40.21 & 37.14 & \quad 38.68 \quad\quad  \\ 
+ DILBERT & 42.99 & 38.20 & \quad 40.60 \quad\quad   \\ 
+ LMLM & \textbf{43.91} & \textbf{39.38} & \quad \bf 41.65 \quad\quad  \\ 

\bottomrule
\end{tabular}
\caption{Results of the RTC dataset. }
\label{tabel:rtc-main-result}
\end{table}


\section{Study}
In this section, we conduct extensive studies to help better understand our method.
It is worth noting that all the experiments in this section are conducted on the BERT model with the ARXIV dataset unless there is a clear explanation.

\subsection{Effect of Pre-training Data Selection}
\label{sec:eff_pretrain_data}
To explore the temporal impact brought by pre-training data, we launch experiments with MLM objective under two different pre-training settings:
\begin{itemize}[leftmargin=*]
    \setlength{\itemsep}{0pt}
    \setlength{\parskip}{0pt}
    \item \textbf{Source Year Consistent Pre-training (SYCP)} We keep the time of pre-training data consistent with that of fine-tuning data to ensure the consistency between the two stages.
    \item \textbf{Target Year Consistent Pre-training (TYCP)} Following the previous work \cite{rottger2021temporal, lazaridou2021mind}, we utilize the pre-training data in consistent with the evaluation data in temporal dimension, i.e., the time of pre-training data and evaluation data is same.
\end{itemize}

We also implement our LMLM objective in the pre-training stage for comparison, where the masking ratio is 15\%, and $k$ is 1000.
The results are shown in the table \ref{tabel:different-pretrain-data}.
We can find that the performance of SYCP gradually overwhelms the TYCP as the time passes towards the target year. 
When the PLM is pre-trained with MLM objective under the SYCP setting, it can even outcome the performance of TYCP, and the PLM pre-trained with LMLM objective under the TYCP setting can achieve the best performance.
On the one hand, we can infer that the temporal adaptation method is effective since TYCP beats the SYCP.
On the other hand, the LMLM objective can make the PLM pay more attention to the salient semantic changed words as pre-training with the LMLM objective under the SYCP settings (SYCP+LMLM) can even surpass the original temporal adaptation method (TYCP).

\begin{table}[t]
\centering
\resizebox{\columnwidth}{!}{
\begin{tabular}{c|cccc|c}
\toprule
\multirow{2}{*}{Method} & \multicolumn{5}{c}{Fine-tuning Data} \\
\cmidrule{2-6}
& 2007 & 2011 & 2015 & 2019 & Avg. \\ 
\midrule
SYCP & 82.81  & 84.62  & 85.54 & 85.17 & 84.54 \\
TYCP (TAda) & 82.97 & 84.72 & 84.82 & 84.99 & 84.38 \\
\midrule
SYCP + LMLM & 83.52 & 85.41 & 86.36 & 87.03 & 85.58 \\
TYCP + LMLM & \textbf{84.93} &  \textbf{86.52} & \textbf{86.49} & \textbf{87.22} & \bf 86.29 \\
\bottomrule
\end{tabular}}
\caption{Results of different pre-training strategies.}
\label{tabel:different-pretrain-data}
\end{table}

\subsection{Hyper-Parameter Analysis}
\label{sec:hyp_LMLM}
\begin{figure}[t]
    \centering
    \includegraphics[width=1.1\linewidth]{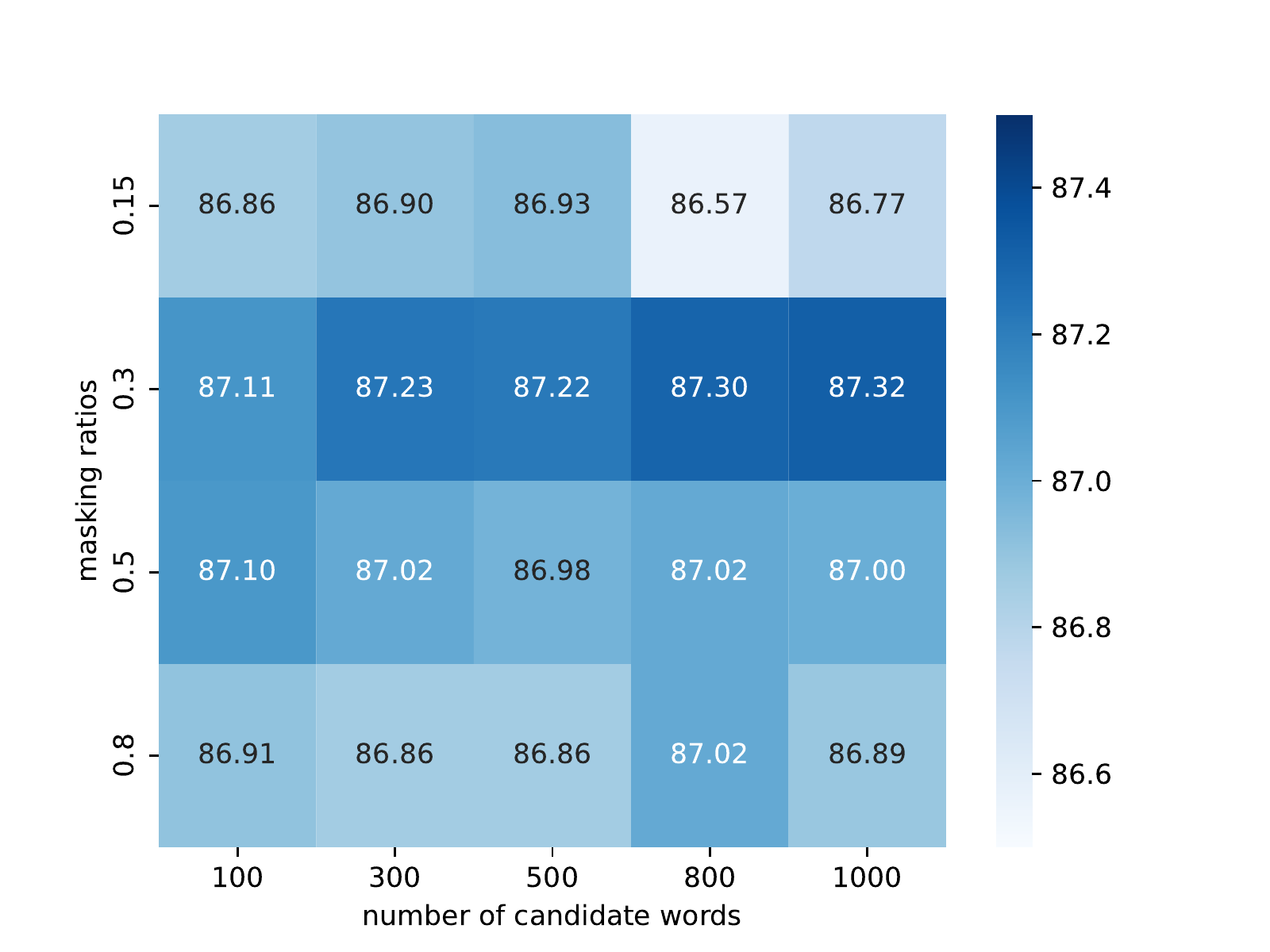}
    \caption{Results of the different masking strategies of LMLM. 
The horizontal axis indicates the number of masked semantic changed words $k$ and the vertical axis represents for the masking ratio $\alpha$.}
    \label{fig:heat}
\end{figure}

Since there is a strong relationship between the masking ratio and the model's performance, we conduct experiments to look for the best masking strategy for the LMLM objective.
Furthermore, we also want to know whether the temporal misalignment problem can be better mitigated by masking more salient semantic changed words. 
Thus, we select the data in 2021 for pre-training with our LMLM objective and the data in 2009 for fine-tuning, followed by testing the model with the data in 2021. 
For better comparison, we utilize a heat map (Figure~\ref{fig:heat}) to display the results, where the vertical axis of this graph represents the masking ratio $\alpha$, and the horizontal axis represents the number of masked salient semantic changed words $k$.

\paragraph{The Influence of $\alpha$}
We calculate the average value of the results of each masking ratio under different settings of $k$ and observe that when the masking ratio is around 30\%, the PLM can achieve the best performance.

\begin{table}[]
\centering
\small
\begin{tabular}{l|lll}
\toprule
JSD~($\uparrow$)\quad\quad  & \#Sen \quad\quad& Prec.\quad\quad & SeC. \\
\midrule
0.00$\sim$0.05 & 902 & 90.2\% & micro \\
0.05$\sim$0.10 & 78  & 7.8\% & medium \\
0.10$\sim$0.15 & 18  & 1.8\% & great \\
0.15$\sim$0.2 & 2  & 0.2\% & great \\
\bottomrule
\end{tabular}
\caption{Distribution of the semantic changed words, where SeC. represents for the \textbf{Se}mantic \textbf{C}hange.}
\label{tab:dis_seman_words}
\end{table}

\paragraph{The Influence of $k$}
No doubt forcing the model to predict more high semantic change words can better mitigate the temporal problem generally.
However, it is surprising to observe that the improvement is slight across different settings of $k$.
Thus, we quantify the semantic change of 1,000 random sampled words from the candidates $\mathcal{W}^{t}_{mask}$ according to the JSD value and the distribution of those words is shown in Table~\ref{tab:dis_seman_words}.
We can find that only around $10\%$ words have relative significant semantic change (JSD value $\geq 0.05$) while around $72\%$ words have little or no semantic change (JSD value $\approx 0.00$).
We can conclude that the improvement mainly comes from predicting a few keywords, i.e., topic words and newly emerging words, which have relatively salient semantic change.

\subsection{Pre-trained Language Model Analysis}
To verify the generalization of our methods on different PLMs, we implement our method on two PLMs, i.e., BERT and RoBERTa, and utilize the temporal adaptation method~\cite{rottger2021temporal} as the baseline for comparison.
As shown in table~\ref{tabel:pre_train_model}, we find there is a dramatic improvement of each PLM, i.e., 1.42 points improvement of the BERT model and 0.48 points improvement of the RoBERTa model on average.

\begin{table}[t]
\centering
\small
\begin{tabular}{c|cccc|c}
\toprule
\multirow{2}{*}{PLMs} & \multicolumn{5}{c}{Fine-tuning Data} \\
\cmidrule{2-6} 
 & 2007 & 2011 & 2015 & 2019 & Avg. \\ 
\midrule
BERT & 82.97 & 84.72 & 84.82 & 84.99 & 84.38 \\
+ TSC-Ada & \textbf{84.93} & \textbf{86.52} & \textbf{86.49} &  \textbf{87.22} & \bf 85.80 \\ 
\midrule
RoBERTa & 81.72 & 84.37 & 84.46 & 84.95 & 83.88 \\ 
+ TSC-Ada & \textbf{82.32} & \textbf{84.92} & \textbf{84.59} & \textbf{85.40} & \bf 84.36 \\
\bottomrule
\end{tabular}
\caption{Results of different PLMs under the time-stratified settings.}
\label{tabel:pre_train_model}
\end{table}
\begin{table}[t]
\centering
\small
\begin{tabular}{c|cccc|c}
\toprule
\multirow{2}{*}{Metrics} & \multicolumn{5}{c}{Fine-tuning Data} \\
\cmidrule{2-6}
& 2007 & 2011 & 2015 & 2019 & Avg. \\ 
\midrule
ED & 84.99 & \textbf{86.62} & 86.43 & \textbf{87.60} & \textbf{86.41} \\
APD & \textbf{85.02} & 86.34 & 86.24 & 87.08 & 86.17 \\
JSD & 84.93 & 86.52 & \textbf{86.49} & 87.22 & 86.29 \\
\bottomrule
\end{tabular}
\caption{Results of different quantification metrics.}
\label{tabel:Comparison_of_metric}
\end{table}
\begin{table}[!tbp]
\centering
\small
\begin{tabular}{l|ccc|c}
\toprule
\multirow{2}{*}{Method} & \multicolumn{4}{c}{Fine-tuning Data} \\
\cmidrule{2-5}
    & \quad 2014  & 2015  & 2016 \quad & Avg. \quad   \\ 
\midrule
 TAda       & \quad 81.23 & 80.91 & 81.94 \quad & 81.36 \quad \\
 + LMLM  & \quad \bf 81.41 & \bf82.50 & \bf82.63 \quad & \bf 82.18 \quad \\ 
\bottomrule
\end{tabular}
\caption{Results of the CoNLL dataset.}
\label{tab:NER}
\vspace{-0.4cm}
\end{table}

\begin{table*}[!tbp]
    \centering
    \small
    \begin{tabular}{l | cc | cc | cc | cc}
    \toprule
    \multirow{2}{*}{\bf Settings} & \multicolumn{2}{c|}{2007} & \multicolumn{2}{c|}{2011} & \multicolumn{2}{c|}{2015} & \multicolumn{2}{c}{2019} \\
     & LMLM & TAda & LMLM & TAda & LMLM & TAda & LMLM & TAda \\
     \midrule
     Results (Total) & \bf 84.32 & 82.27 & \bf86.28 & 84.60 & \bf86.17 & 85.13 & \bf87.17 & 85.18 \\
     Results (w/o Temp) & \bf 83.81 & 81.80 & \bf85.99 & 83.84 & \bf86.38 & 85.68 & \bf87.60 & 85.99 \\
     Results (w/ Temp) & \bf 84.02 & 82.79 & \bf86.40 & 84.97 & \bf85.62 & 84.30 & \bf86.66 & 84.35 \\
     \midrule
     Mask (Failed Sets) & 12.87 & \bf 12.91 & 10.75 & \bf11.25 & 8.95 & \bf9.96 & 10.41 & \bf13.04 \\
     PAD (Failed Sets) & 12.62 & \bf 13.34 & 11.15 & \bf11.02 & 8.81 & \bf10.14 & 10.58 & \bf12.77 \\
     REP (Failed Sets) & \bf 13.58 & 13.23 & 10.89 & \bf11.75 & 8.81 & \bf9.78 & 9.92 & \bf13.57 \\
     \bottomrule
    \end{tabular}
    \caption{Error analysis of the LMLM method, where the first group shows the results on the hierarchical data (w/ and w/o temporal information) while the second group shows the results on the failed examples.}
    \label{tab:error_analysis}
\end{table*}

\subsection{Quantification Metrics}
As mentioned above, there are several metrics to quantify the semantic change, and we primarily conduct the experiment to select the JSD metric 
and we compare three commonly used metrics, i.e., ED, APD, and JSD, in this section.
As shown in the table~\ref{tabel:Comparison_of_metric}, we can find that although different metrics have their advantages, the differences among them are slight.
For example, the maximum difference is 0.24 points among three metrics on average.

\subsection{Open-Domain Temporal Adaptation}
As mentioned above, we conduct all the experiments under the domain-specific setting.
In this section, we explore the effect of the LMLM objective with the name entity recognition task under the open-domain setting, i.e., the downstream dataset has no specific domain.
Specifically, we select the WMT dataset in 2017 as the unlabeled data and the subset in 2015 and 2016 from the CoNLL dataset\footnote{\url{https://github.com/shrutirij/temporal-twitter-corpus}} as the fine-tuning data.
In the end, we evaluate the model with the data in 2017.
The results are shown in Table~\ref{tab:NER}.
We can find that the LMLM method outperforms the original temporal adaptation method with around 1 point improvement.

\subsection{Error Analysis}
We also conduct fine-grained experiments to study why our method fails with some examples.
Specifically, we utilize the ARXIV dataset from 2007 to 2017 to fine-tune the model and the data in 2021 for testing.
We first select the top 100 lexical semantic changed tokens for each testing set.
Then, we divide the testing data into two parts: a subset with temporal information and a subset without temporal information by judging whether the texts contain the selected tokens.
As shown in the first group of Table~\ref{tab:error_analysis}, the LMLM method can achieve better results than TAda on both testing subsets, and the improvement on the subset with temporal information is more significant than that on the subset without the  temporal information.
A possible explanation for why LMLM performs better on the subset without temporal information is that there is still some temporal information left in this data since we distinguish the subset with only 100 lexical semantic changed tokens.

Besides, we collect the failed testing sets, i.e., the model predicting wrong labels on those data, and mask those mentioned above top 100 lexical semantic changed tokens in the texts with two strategies: (1) replace those tokens with special placeholder \texttt{<MASK>} or \texttt{<PAD>}, and (2) randomly utilize other tokens in the vocabulary (except the aforementioned lexical semantic changed tokens) for substitution.
The results are shown in the second group of Table~\ref{tab:error_analysis}, where we can observe that TAda surpasses our method in general\footnote{It is worth noting that LMLM surpasses the TAda on the REP testing set in 2007, which can be attributed to the possibility of replacing the original tokens with lexical semantic changed tokens.}.
We think those masked/replaced lexical semantic changed tokens, which LMLM pays more attention to, may be the critical messages for the model to help the decision.
The missing of that important information can cause a negative impact on the model, which leads to the performance decreasing.
\section{Related Work}
\subsection{Temporal Misalignment}
Previous studies have shown that models trained on texts from one time period perform poorly when tested on texts in later periods for NLP tasks like machine translation~\cite{levenberg2010stream}, review and news article classification~\cite{huang2019neural,huang2018examining}, named entity recognition~\cite{rijhwani2020temporally} and so on.
Within the current paradigm of using PLMs~\cite{devlin2019bert}, studies have focused more on the expansion of dataset~\cite{liu2019roberta, lewis2020bart, yang2019xlnet} and model capacity~\cite{raffel2019exploring, lan2019albert, brown2020language} to achieve better performance but ignore the temporal effects.
Few studies focus on such problem, \citet{lazaridou2021mind} have empirically studied the degraded performance of PLMs over time, and \citet{rottger2021temporal} focus on post-tuning BERT with the data in specific periods to mitigate the temporal misalignment problems.
Furthermore, \citet{amba2021dynamic} propose sampling methods to help PLMs achieve better performance on the evolving content.
In this paper, we conduct a detailed investigation from the perspective of lexical semantic change to figure out the reason behind the limited performance of the PLMs under the time-stratified settings. 


\subsection{Lexical Semantic Change}
Lexical semantic change is an extensively studied concept in the computational linguistics, which mainly focuses on deciding whether the concept of a word has changed over time (semantic change detection)~\cite{gulordava2011distributional,kulkarni2015statistically,dubossarsky2015bottom,hamilton2016diachronic} or discovering the instances with high semantic change (semantic change discovery)~\cite{hengchen2021data,kurtyigit2021lexical,jatowta2021computational}.
Among them, most studies utilize contextualized word representations~\cite{turney2010frequency,giulianelli2020analysing} and measure the distance among them in different periods~\cite{cook2010automatically, gulordava2011distributional, hamilton2016diachronic} to detect or discover the instances with salient semantic change.
Previous studies mainly concentrate on applying PLMs to discover the semantic change phenomena, while our work focuses on solving such problems intrinsic in the PLMs.
Thus, besides observing such semantic changed phenomenon, we aim to find the corresponding words and apply the LMLM objective to make the PLMs more aware of them to mitigate the temporal misalignment problem.
Most studies focused on obtaining those words are under the supervised settings~\cite{kim2014temporal,basile2016diachronic,basile2018exploiting,tsakalidis2019mining} by scoring and selecting the top-ranked words through author intuitions or known historical data~\cite{kurtyigit2021lexical}.
While \citeauthor{giulianelli2020analysing} propose one unsupervised method, adding one clustering process to the traditional selecting methods.
To our best knowledge, this is the first work that links semantic change with temporal adaptation.

\section{Conclusion \& Future Work}
In this paper, we investigate the temporal misalignment of the PLMs from the lexical level and observe that the words with salient lexical semantic change contribute significantly to the temporal problems.
We propose a lexical-based masked Language Model (LMLM) objective based on the above observation.
Experiments on two PLMs with the sequence classification task on three datasets under the specific domain setting and one name entity recognition task under the open-domain setting confirm that our proposed method performs better than the previous temporal adaptation methods and the state-of-the-art domain adaptation methods.
In the future, we will keep discovering such temporal misalignment problems in the text generation tasks, e.g., machine translation, and improve our method by reducing the extra offline computational cost on procedures like Semantic Change Detection.

\section{Limitation}
There are still some limitations in our work which are listed below:
\begin{itemize}[leftmargin=*]
    \setlength{\itemsep}{0pt}
    \setlength{\parskip}{0pt}
    \item The other tokens in the text influence the meaning of the target word to some extent since we utilize sentence contextualization to represent the meaning of the target word. 
    To this end, it is hard to interpret why some candidate words are selected by the detection step, e.g., name entities or numbers, whose meaning remains unchanged.
    We will design a better unsupervised word selection strategy in the future. 
    \item We utilize the lexical-level masking strategy, while the semantic change can also be reflected with the whole sequence, e.g., the topic of ``Malaysia Airlines crashed into the sea'' may be one hypothesis before 2014, but it became a severe accident in 2014.
    Current famous MLM objectives like span masking objective~\cite{raffel2020exploring} or sentence masking objective~\cite{tay2022unifying} have observed that the performance of denoising the whole sequence is better than denoising the single token in some NLU tasks.
    In the future, we will explore whether the sequence masking objective mentioned above can mitigate the temporal misalignment problem inherent in the PLMs.
\end{itemize}

\section*{Acknowledgement}
This work was supported by the National Science Foundation of China (NSFC No. 62206194), the Natural Science Foundation of Jiangsu Province, China (Grant No. BK20220488), and the Project Funded by the Priority Academic Program Development of Jiangsu Higher Education Institutions.

\bibliography{anthology}
\bibliographystyle{acl_natbib}

\clearpage
\appendix
\section{Implementation of the K-Means Algorithm}
\label{appdix:k_means}
Given a representation matrix $\mathcal{R}^{t}_{i} = \{r^{t}_{i}\}_{i\in (1, \cdots, N)}$ for the word $w^{t}_{i}$ we utilize the silhouette score~\cite{rousseeuw1987silhouettes} to obtain the optimal $K$ for K-Means algorithm.
We experiment with $K \in [2,10]$ in a heuristic way.
For each $K$, the clustering result is the one that yields the minimal distortion value, i.e., the minimal sum of squared distances of each data point from its closest centroid, and we execute ten iterations to alleviate the influence of different initialization values~\cite{vassilvitskii2006k}.
Since there are several monosemous words, i.e., the number of $K$ is 1, we filter those words with a threshold $d$.
Specifically, if the intra-cluster dispersion value of a word is below $d$, we would allocate $K = 1$, otherwise, $K \geq 2$.
The optimal $K$ is the one that can simultaneously minimize the dispersion score and maximize the silhouette score.

\section{Implementation of the Masking Strategies}
\label{appdix:semantic}
In this section, we illustrate the frequency masking strategy and importance masking strategy in detail.
Given a dataset that contains $n$ texts, we firstly utilize the NLTK tool\footnote{\url{https://github.com/nltk/nltk}} to tokenize each text and follow the below processes:
\paragraph{Frequency Masking Strategy}
We add each tokenized token into the dictionary $\mathcal{D}$ and record the number of its occurrence.
We sort the tokens in $\mathcal{D}$ according to the occurrence times and select the tokens to mask in descending order until the masking ratio is satisfied.

\paragraph{Importance Masking Strategy}
We utilize the YAKE! method as mentioned above to sort the tokenized tokens according to the scores calculated with the task label, e.g., the label of CS in the ARXIV dataset.
Finally, we select the tokens to mask in descending order until the masking ratio is satisfied.

\section{Model Training \& Parameters}
\label{appdix:para}
\paragraph{Architecture}
We utilize the BERT-base uncased model pretrained on a large corpus of English data with the MLM objective.
The model contains 12 transformer layers, 12 attention heads, and the hidden layer size is 768.
The total number of parameters is 110 million.
We add a linear layer after the last BERT layer for the downstream classification task and generate the output with softmax.
The maximum input sequence length is 512.

\paragraph{Training Details}
We utilize cross-entropy loss in the pre-training and fine-tuning stages and apply AdamW~\cite{loshchilov2017decoupled} as the optimizer.
Specifically, the learning rate is 5e-5, and the weight decay is 0.01.
Moreover, we set a 10\% dropout probability for regularisation,
We pre-train the model for one epoch and fine-tune the model until convergency.
We set the batch size as 128 and conduct the experiments on eight NVIDIA GTX3090 GPUs.

\paragraph{Evaluation Metric}
We utilize the F1 score\footnote{\url{https://scikit-learn.org/stable/modules/generated/sklearn.metrics.f1_score.html}} as the evaluation metric in all the experiments.

\section{Implementation of the Baselines}
\label{appdix:methods}
This section will elaborate on how to apply the domain adaptation methods under the temporal adaptation settings.
\paragraph{PERL} ~\cite{ben2020perl} This method  model
parameters using a pivot-based ~\cite{blitzer2006domain, blitzer2007biographies} variant of the MLM object with unlabeled datasets from both the source and target temporal split. 
Instead of masking each token with the same probability, we divided token into pivots and non-pivots to learn the pivot/non-pivot distinction on unlabeled data from the source and target time span. The encoder weights are frozen during training for the downstream task.
Specifically, we rank those frequent features (occurs at least 20 times in the unlabeled data from the source and target time split) based on the mutual information with the task label according to source domain labeled data.
Then, we select top 100 which have relative high scores as pivot features.
The non-pivot feature subset consists of features that do not match the two requirements.

\paragraph{DILBERT} 
~\cite{lekhtman2021dilbert} is the SOTA in Aspect Extraction while using a fraction of the unlabeled data.
Different from PERL, they challenge the “high MI with the task label" criterion in the pivot definition. In our settings, we harness the information about the golden label(physics, cs., etc) in the source and target temporal split to mask words that are more likely to bridge the gap between the different periods. Specifically, we compute the cosine similarity between each input word and the label from both the source and the target. We keep the highest similarity score for each word and mask the top 0.15\% of the input words. For the downstream task, they add a logistic regression head on top of all outputs and fine-tune the model on the source period labeled data.

\section{Feature Extracting}
\label{appdix:upper_layer}
One point that worth discussing is the hidden states from the last layer of the BERT model ($\mathrm{LMLM_{\texttt{LAST}}}$) contains massive amounts of contextual information, which may overwhelm the lexical information.
Thus we turn to the representation from the shallow layer of the BERT model, e.g., representation from the second BERT layer ( $\mathrm{LMLM_{\texttt{SECOND}}}$).
Specifically, we conduct the experiment on the ARXIV testing set in 2013, and the results are shown in Table~\ref{tab:rep_layer}.

\begin{table}[h]
    \centering
    \begin{tabular}{l|l}
    \toprule
        Model & F1 \\
    \midrule
        $\mathrm{LMLM_{\texttt{LAST}}}$ & \bf 87.45 \\
        $\mathrm{LMLM_{\texttt{SECOND}}}$ & 86.67 \\
        TAda & 85.04 \\
    \bottomrule
    \end{tabular}
    \caption{Results on the ARXIV testing set.}
    \label{tab:rep_layer}
\end{table} 

As we can observe from the table that the $\mathrm{LMLM_{\texttt{SECOND}}}$ can achieve a better result than TAda, which indicates that the representations from the earlier layer are strong enough to help achieve a decent performance improvement.
Besides, the $\mathrm{LMLM_{\texttt{LAST}}}$ achieves a better result than $\mathrm{LMLM_{\texttt{SECOND}}}$, which means that the hidden states which contain sentence-level information can help promote the accuracy in the detection process.







\end{document}